\documentclass{IEEEcsmag}

\usepackage[colorlinks,urlcolor=blue,linkcolor=blue,citecolor=blue]{hyperref}
\expandafter\def\expandafter\UrlBreaks\expandafter{\UrlBreaks\do\/\do\*\do\-\do\~\do\'\do\"\do\-}
\usepackage{upmath,color}

\usepackage{url}
\usepackage{amsmath}
\usepackage{graphicx}
\usepackage{amsfonts}
\usepackage{bm}
\usepackage{subcaption}
\graphicspath{{figs/}}
\usepackage{comment}
\usepackage{tikz}
\def\checkmark{\tikz\fill[scale=0.4](0,.35) -- (.25,0) -- (1,.7) -- (.25,.15) -- cycle;}

\jvol{XX}
\jnum{XX}
\paper{8}
\jmonth{Month}
\jname{Publication Name}
\jtitle{Publication Title}
\pubyear{2021}

\newcommand\blfootnote[1]{%
  \begingroup
  \renewcommand\thefootnote{}\footnote{#1}%
  \addtocounter{footnote}{-1}%
  \endgroup
}

\begin{document}

\sptitle{Department Article: Machine Learning for Dynamic Envirnoments}

\title{Streaming Continual Learning for Unified Adaptive Intelligence in Dynamic Environments}

\author{Federico Giannini, and Giacomo Ziffer}
\affil{DEIB, Politecnico di Milano, Milano, 20133, Italy}

\author{Andrea Cossu, and Vincenzo Lomonaco}
\affil{Computer Science Department, University of Pisa, Pisa, Italy}

\markboth{DEPARTMENT}{DEPARTMENT}

\begin{abstract}\looseness-1
Developing effective predictive models becomes challenging in dynamic environments that continuously produce data and constantly change. Continual Learning (CL) and Streaming Machine Learning (SML) are two research areas that tackle this arduous task.
We put forward a unified setting that harnesses the benefits of both CL and SML: their ability to quickly adapt to non-stationary data streams without forgetting previous knowledge. We refer to this setting as Streaming Continual Learning (SCL). SCL does not replace either CL or SML. Instead, it extends the techniques and approaches considered by both fields. 
We start by briefly describing CL and SML and unifying the languages of the two frameworks. We then present the key features of SCL. We finally highlight the importance of bridging the two communities to advance the field of intelligent systems.
\end{abstract}

\maketitle
\blfootnote{\copyright~2024 IEEE. This is the author’s accepted manuscript of the article: Giannini, F., Ziffer, G., Cossu, A., Lomonaco, V. (2024). \textbf{Streaming Continual Learning for Unified Adaptive Intelligence in Dynamic Environments}. \textit{IEEE Intelligent Systems}, 39(6), 81–85. The final authenticated version is available online at \url{https://doi.org/10.1109/MIS.2024.3479469}}

\chapteri{T}he world we experience as humans is dynamic, constantly inundated with new information altering our beliefs, plans, and actions. Similarly, our environment changes. Developing robust prediction models in such dynamic settings remains a challenge. Machine Learning (ML) struggles in non-stationary environments where data evolves over time, excelling only in static conditions.

To learn in dynamic environments, Continual Learning (CL)~\cite{lesort2020} and Streaming Machine Learning (SML)~\cite{book_bifet} research areas have emerged. Although they share similarities, they have rarely interacted. They leverage comparable definitions of ``data streams'', focusing on \emph{dynamic} data streams, where the presence of drifts requires the learning model to adapt to new trends. Their main distinction lies in the objectives. The goal of CL is to \emph{learn new information over time from an evolving data stream without forgetting previously acquired knowledge}~\cite{mccloskey1989}. Forgetting is a well-known phenomenon in CL that causes a deterioration of the performance on previous data when learning new information. Conversely, the main goals of SML are \emph{quick adaptation to drifts and real-time predictions}. SML usually ignores the problem of forgetting and adapts as quickly as possible to the current data. CL also builds hierarchical representations using Deep Learning models. Instead, SML is subject to strict temporal and memory constraints and usually produces streaming versions of Statistical ML models. The paradigm of Online Continual Learning (OCL) \cite{mai2022}  emerged to blend CL and SML perspectives. However, OCL is heavily focused on the CL objectives. A survey by Gunasekara et al.~\cite{gunasekara2023} explores the concept of "Online Streaming Continual Learning," suggesting ways CL and SML could complement each other. Precisely, its main contribution regards the usage of SML techniques to automatically detect drifts, which is usually absent in CL. While it does not propose a unified framework, its insights offer valuable directions. Conversely, we see a growing need for a balanced approach merging CL and SML goals, which we will call Streaming Continual Learning (SCL). This unified framework will not replace SML or CL but will handle scenarios they, alone, cannot. Our SCL framework draws from the Complementary Learning Systems (CLS) theory~\cite{kumaranWhatLearningSystems2016,mcclellandIntegrationNewInformation2020}, describing the human decision-making process that blends fast and slow learning. In our SCL framework, the CL agent retains knowledge over time while the SML agent rapidly adapts to new data.

\begin{figure*}[t]
    \centering
    \begin{subfigure}{0.45\textwidth}
        \includegraphics[width=0.95\textwidth]{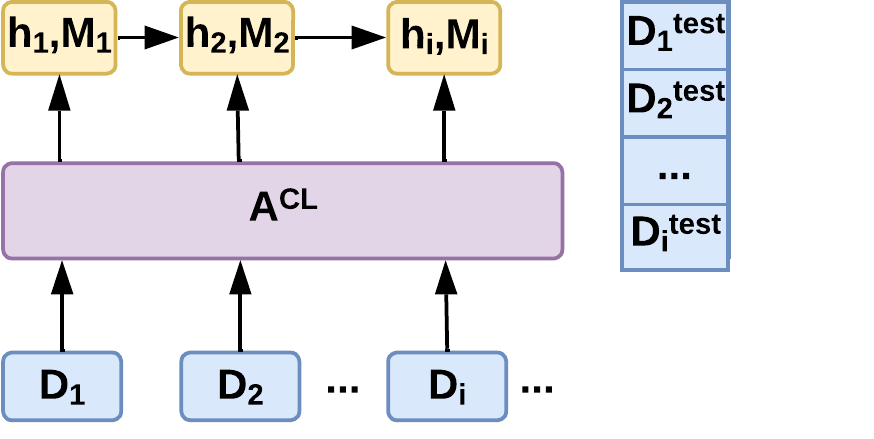}
        \caption{}
        \label{fig:cl}
    \end{subfigure}
    \hfill
    \begin{subfigure}{0.45\textwidth}
        \includegraphics[width=0.95\textwidth]{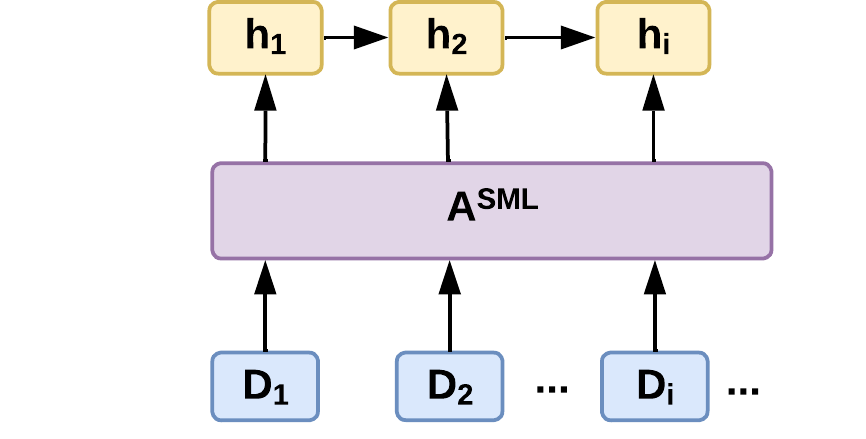}
        \caption{}
        \label{fig:sml}
    \end{subfigure}
    \caption{Comparison between a Continual Learning algorithm $A^{\text{CL}}$ and Streaming Machine Learning algorithm $A^{\text{SML}}$ trained on a possibly infinite sequence of items, indexed by the subscript $i$. (\subref{fig:cl}) Continual Learning with a model $h$ and an external memory $M$. Training is performed on one training item's data $D_i$ at a time. The test sets $D_1^{\text{test}}, D_2^{\text{test}}, \ldots, D_i^{\text{test}}$ are used for evaluation. (\subref{fig:sml}) Streaming Machine Learning with a model $h$ and a stream of items' data $D_i$, where each $D_i$ contains a mini-batch of a few examples. Each $D_i$ is used both for the testing and the training phases.}
    \label{fig:cl-sml}
\end{figure*}

In this paper, after briefly describing CL and SML, we examine the key properties required by a unifying SCL scenario. We also discuss future research directions to unite CL and SML communities and promote adaptive intelligence in dynamic environments.

\section{NON-STATIONARY DATA STREAMS}

Despite notable distinctions between CL and SML, they address similar non-stationary environments. The shared terminology of "data stream" and "drift" offers a common ground for fostering collaboration between CL and SML research. This is precisely the opportunity we envision with Streaming Continual Learning.

A data stream can be defined as a potentially infinite and ordered sequence of items $(e_1, e_2, e_3, \ldots)$ \cite{lomonaco2021}. Crucially, the learning model $h$ is provided with one item at a time and cannot access the entire stream simultaneously. As shown in Figure~\ref{fig:cl-sml}, each item carries a dataset $D_i$ containing a set of K examples. In the case of supervised problems, $D_i$ contains a set of input-target pairs ${(\bm{X_j}, \bm{y_j})}_{j=1,\ldots,K}$, where $\bm{X_j} \in \mathbb{R}^d$ is an input vector with $d$ features and $\bm{y_j} \in \mathbb{R}^c$ is the target vector with $c$ features\footnote{$c$ can be the number of classes in a classification task.}. A crucial difference between CL and SML data streams is how each item is constructed. In CL, items are usually called \emph{experiences} that contain many examples. The CL learning model can visit each example for as long as it takes to acquire the new information. It can be evaluated anytime on test datasets (one for each experience). Instead, each item in SML carries only one example (or a very small number). Therefore, the model must quickly learn new information in a \emph{single pass}. The model is first evaluated on the examples from the incoming experience, and then the same examples are used to train the model (\emph{prequential} evaluation~\cite{gama2009evaluation}). A CL algorithm $A^{CL}$~\cite{lesort2020} and an SML algorithm $A^{SML}$ are very similar (Figure~\ref{fig:cl-sml}). Both take as input one item $e_i$ at a time, and, given the current learning model $h_i$, they output the next model $h_{i+1}$. 
A CL algorithm sometimes leverages an external, fixed-size memory to store additional information, like previously encountered examples.

\begin{figure*}[t]
    \centering
    \includegraphics[width=\textwidth]{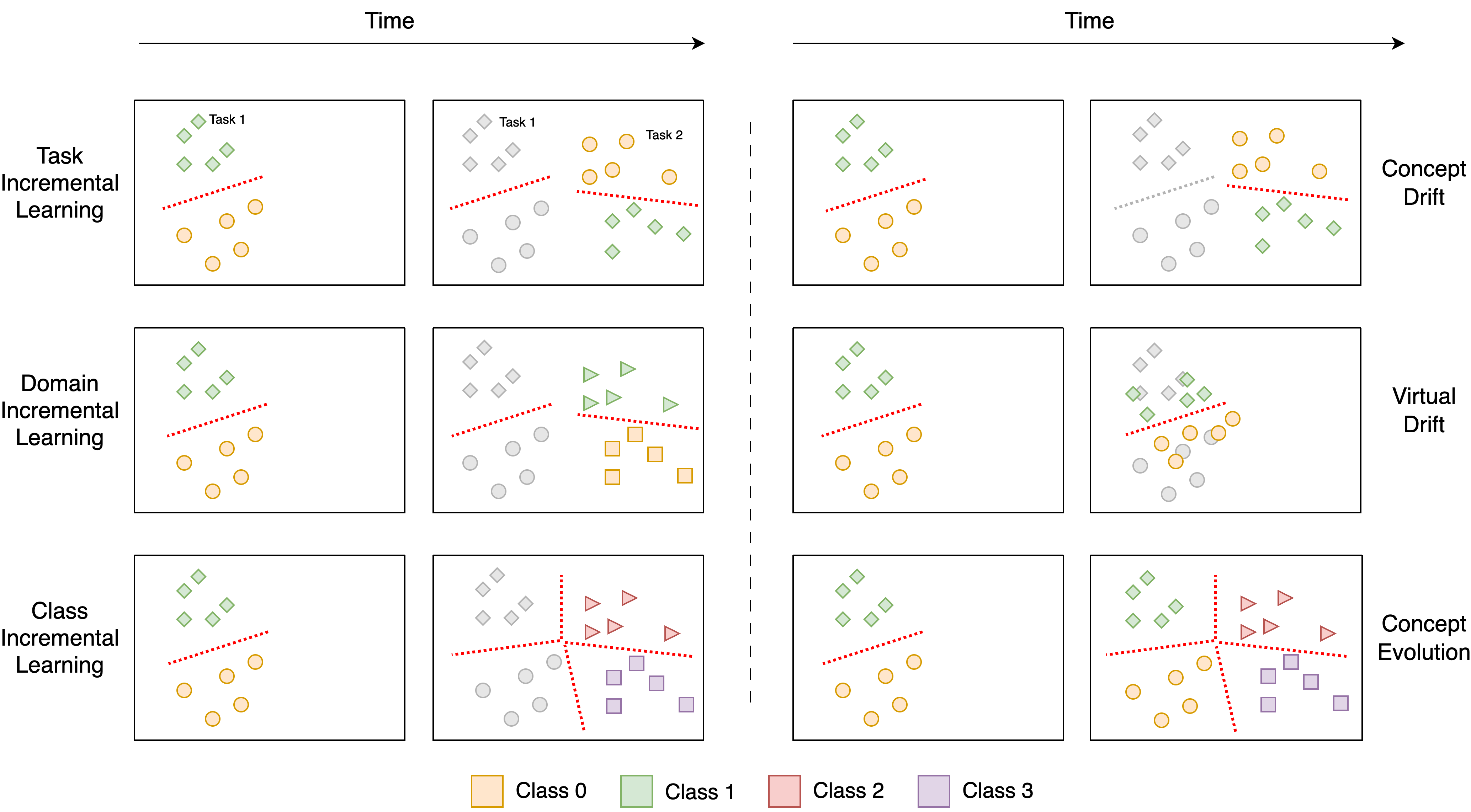}
    \caption{Comparison between Continual Learning Scenarios and Drifts in Streaming Machine Learning. The red dashed line highlights the decision boundary, while grey dots represent data no longer available. Different shapes represent novel data that may still fall into the same classes as the previous experience, as per Domain Incremental Learning, or may represent new classes (denoted by different colors), as per Class Incremental Learning. In Domain-Incremental learning, each experience contains examples from all classes, while new experiences introduce new examples only, potentially leading to virtual drift. Class-Incremental learning is prone to concept evolution since it exclusively introduces examples of new classes in each new experience. Task-Incremental learning assigns a task label to each example. It can be viewed as a concept drift or a concept evolution, depending on whether we consider the class labels of all tasks or only those of the current task.}
    \label{fig:data}
\end{figure*}

A central property of CL and SML is their ability to deal with \emph{concept drifts}, which are unforeseeable changes in the data-generating process that induce a corresponding change in the data statistical properties. This change is more significant than one caused by random fluctuations or anomalies~\cite{CD_SURVEY1}. \emph{Virtual drifts} occur when probabilities $P(X|y)$ or $P(y)$ change, not affecting the class boundary. \emph{Real drifts} occur with a change in $P(y|X)$, altering the class boundary. Moreover, a \emph{concept evolution} models the case where new class labels appear over time. Drifts can occur suddenly (\emph{abrupt drift}) or gradually (\emph{gradual drifts}). Additionally, a \emph{recurring scenario} reintroduces old concepts in the future. CL assumes each new experience induces an abrupt drift, while SML focuses on detecting drift occurrences. In the case of CL for supervised classification tasks, there are three main types of data streams. In a Domain-Incremental scenario, all classes are present within each experience, allowing for the gradual refinement of concepts. Conversely, Class-Incremental scenario introduces new classes with each experience, leading to the evolution of decision boundaries. Finally, Task-Incremental learning employs unique labels to aid in task separation, yet real-world limitations constrain its utility. In Figure~\ref{fig:data}, we provide a correspondence between these three scenarios for CL and the SML drifts discussed before. Although a one-to-one mapping is impossible, we try to see them through the lenses of drifts.

\section{STREAMING CONTINUAL LEARNING} \label{sec:unified}
The paradigm of OCL has recently emerged as a first step towards integrating CL and SML. In OCL, each experience carries only a few examples, like in SML. 
However, OCL remains mostly focused on the CL objectives without integrating SML ones. For simplicity, we will use ``CL'' to refer to both CL and OCL paradigms.
We aim to propose an effective approach to merge CL and SML within a unified paradigm that we call \emph{Streaming Continual Learning} (SCL). SCL has the six key properties summarized in Table~\ref{table:sml_cl_scenarios}.

\subsection{Quick Adaptation Without Forgetting}
Adopting the SCL paradigm requires the model to learn from a limited number of examples while remaining \emph{oblivious} to incoming drifts. SML provides a robust way to \emph{evaluate} the ability of a model to adapt to incoming drifts with prequential evaluation. However, this is insufficient since it does not measure forgetting. In SCL, we start from the assumption that it is \emph{the environment that dictates what is important and what can be forgotten.}  For instance, in a concept evolution scenario, new experiences can introduce new classes needing integration without forgetting previous ones. In other cases, preserving concepts that will not reappear is not efficient. Current CL practices often retain all experiences in test sets, even those no longer relevant. In Class/Task-Incremental scenarios, previous classes will not return, so forgetting can aid adaptation to new data. We suggest using the prequential evaluation to assess adaptation to drifts. External test sets can monitor performance on selected concepts, while CL techniques can mitigate forgetting of relevant ones. Monitoring forgetting helps understand the impact of new information. Even if a concept becomes unimportant, studying its accuracy changes can be insightful. The model should retain shared general knowledge among concepts, which is beneficial for addressing current issues.

\begin{table}
    \centering
    \renewcommand{\arraystretch}{1.2}
    \caption{Comparison of key properties of Continual Learning (CL), Streaming Machine Learning (SML), and the proposed Streaming Continual Learning (SCL) frameworks. We refer to CL for both CL and OCL paradigms. Only a few recent pioneering OCL works apply autonomous drift detection, which is not present in CL. CL does not apply to learn from a few data points, while OCL was meant to fulfil this need. SCL acts as a real unifying paradigm incorporating SML and CL goals and techniques.}
    \label{table:sml_cl_scenarios}
   \begin{tabular}{rccc}
\multicolumn{1}{r|}{}                                                                                    & \multicolumn{1}{c|}{\textbf{CL}} & \multicolumn{1}{c|}{\textbf{SML}} & \multicolumn{1}{c|}{\textbf{SCL}} \\ \hline
\multicolumn{1}{r|}{Quick adaptation to drifts}                                                          & \multicolumn{1}{c|}{}            & \multicolumn{1}{c|}{\checkmark}        & \multicolumn{1}{c|}{\checkmark}        \\ \hline
\multicolumn{1}{r|}{Autonomous drift detection}                                                          & \multicolumn{1}{c|}{\checkmark*}      & \multicolumn{1}{c|}{\checkmark}        & \multicolumn{1}{c|}{\checkmark}        \\ \hline
\multicolumn{1}{r|}{\begin{tabular}[c]{@{}r@{}}Learning from single (few) \\ data points\end{tabular}}   & \multicolumn{1}{c|}{\checkmark*}      & \multicolumn{1}{c|}{\checkmark}        & \multicolumn{1}{c|}{\checkmark}        \\ \hline
\multicolumn{1}{r|}{\begin{tabular}[c]{@{}r@{}}Learning hierarchical\\ representations\end{tabular}}     & \multicolumn{1}{c|}{\checkmark}       & \multicolumn{1}{c|}{}             & \multicolumn{1}{c|}{\checkmark}        \\ \hline
\multicolumn{1}{r|}{\begin{tabular}[c]{@{}r@{}}Catastrophic forgetting\\ avoidance\end{tabular}}         & \multicolumn{1}{c|}{\checkmark}       & \multicolumn{1}{c|}{}             & \multicolumn{1}{c|}{\checkmark}        \\ \hline
\multicolumn{1}{r|}{\begin{tabular}[c]{@{}r@{}}Interaction of CL and SML\\ learning models\end{tabular}} & \multicolumn{1}{c|}{}            & \multicolumn{1}{c|}{}             & \multicolumn{1}{c|}{\checkmark}        \\ \hline
\multicolumn{4}{c}{* Applies to the OCL paradigm only.}                                                                                                                                                            
\end{tabular}
\end{table}

\subsection{A Dual Learning Approach}
While CL builds hierarchical representations on large data, SML is constrained to a fast optimization phase. Moreover, as SML commonly leverages Statistical ML algorithms, it converges with less data than the Deep Learning models employed in CL strategies. In the context of SCL, these properties can be effectively utilized following the CLS theory. This theory argues that humans leverage a fast-learning component (usually associated with the hippocampus) that quickly incorporates new information and a slow-learning component (the neo-cortex) that consolidates the knowledge of the fast-learning component over time. In our vision, an SML model represents a fast learning process focusing on the current concept. A CL agent implements a slow learning process, consolidating knowledge associated with the entire data stream. We, therefore, imagine the fast optimization process of SML to be the high-frequency learner. Data arrive quickly, and the algorithm needs to get the most out of it immediately. As the data stream continues to provide new experiences, the knowledge contained in the fast SML model may inform the slower CL algorithm. Important concepts can only be defined after a certain time and can be preserved using CL techniques. Importantly, this does not come with a trade-off with performance on the incoming data since the SML algorithm would still prioritize prequential accuracy through prequential evaluation. The interaction between fast and slow learners can be bi-directional. The fast SML model can gain insights from the comprehensive knowledge of the slow CL model. As a Deep Learning model, the CL agent computes latent representations encapsulating common patterns and relevant information across all seen concepts. The learned latent representations can serve as a foundation for training the SML model on new concepts. In this way, the fast learner does not learn from scratch but leverages consolidated knowledge to build new concepts quickly. The dual learning approach could also be effective in different practical applications, e.g.,  spatiotemporal data. In time-series forecasting, in fact, the combination of ML and Deep Learning has been proven efficient. Additionally, in cybersecurity, one could examine CL to learn the node embeddings within an authentication network, while SML could be considered to classify anomalous paths. Finally, recognizing when drifts occur is essential to react quickly. SML drift detectors can be applied to the fast learning component.

\section{CONCLUSION}
In this article, we envisioned Streaming Continual Learning, a framework combining the properties of Continual Learning and Streaming Machine Learning when learning from real-time dynamic data streams. Inspired by a popular neuroscientific theory describing human decision-making, we showed how CL and SML can play different but complementary roles in the definition of SCL. SML techniques enable the detection of drifts and a quick adaptation to new information. CL ensures learning new information \emph{while retaining past knowledge}, thus allowing the system to generalize effectively. Combining both approaches gives rise to a Slow System that captures the long-term dependencies and encodes stable knowledge and a Fast System that rapidly adapts to new information. This integration holds great promise for advancing the field of intelligent systems and paves the way for further research and development in lifelong learning and dynamic environments.

\section{ACKNOWLEDGMENTS}
We would like to thank Alessio Bernardo for his expert insights, for assisting in the ideation of the framework, and for highlighting key aspects of Streaming Machine Learning and Continual Learning.

We also thank Prof. Emanuele Della Valle and Davide Bacciu for their supervision while writing this paper and for the stimulating discussions that greatly enriched the work.

\def\refname{REFERENCES}

\bibliographystyle{naturemag}
\bibliography{bibliography, cl}

\begin{IEEEbiography}{Federico Giannini}{\,} is a Ph.D. student at DEIB, Politecnico di Milano, Italy. His research interests lie at the intersection of Streaming Machine Learning and Continual Learning with the goal of jointly addressing the problems of concept drifts, temporal dependence, and catastrophic forgetting. Contact him at \mbox{federico.giannini@polimi.it}.
\end{IEEEbiography}

\begin{IEEEbiography}{Andrea Cossu}{\,} is an Assistant Professor at the Department of Computer Science of the University of Pisa. His research interests revolve around continual and lifelong learning, with applications to recurrent neural networks, sequential data processing and pre-trained models. Andrea received his Ph.D. in Data Science from Scuola Normale Superiore, Italy. Contact him at \mbox{andrea.cossu@unipi.it}.
\end{IEEEbiography}

\begin{IEEEbiography}{Giacomo Ziffer}{\,} is a Ph.D. student at DEIB, Politecnico di Milano, Italy, where he investigates the impact of temporal dependence and concept drift on learning from data streams and time series. His research interests include Streaming Machine Learning, Time-Evolving Analytics, Continuous Time Series Analysis, and Streaming Edge AI. Contact him at \mbox{giacomo.ziffer@polimi.it}.
\end{IEEEbiography}

\begin{IEEEbiography}{Vincenzo Lomonaco}{\,} is a Researcher at the University of Pisa. He also serves as Co-Founding President and Lab Director at ContinualAI, a non-profit research organization and the largest open community on Continual Learning for AI, Co-founding Board Member at AI for People, and as proud member of the European Lab for Learning and Intelligent Systems (ELLIS). In Pisa, he works within the Pervasive AI Lab and the Computational Intelligence and Machine Learning Group. Previously, he was a Post-Doc at University of Bologna where he also obtained his PhD in early 2019 with a dissertation titled ``Continual Learning with Deep Architectures" which was recognized as one of the top-5 AI dissertations of 2019 by the Italian Association for Artificial Intelligence. In the past Vincenzo has been a Visiting Research Scientist at AI Labs in 2020, at Numenta in 2019, at ENSTA ParisTech in 2018 and at Purdue University in 2017. His main research interest and passion is about Continual Learning, in particular for Deep Learning and Distributed Learning within a sustainability developmental framework. Contact him at \mbox{vincenzo.lomonaco@unipi.it}.
\end{IEEEbiography}

\end{document}